\documentclass[electronic]{vgtc}                % final (conference style)
\ifpdf%                                % if we use pdflatex
  \pdfoutput=1\relax                   % create PDFs from pdfLaTeX
  \usepackage{graphicx}                % allow us to embed graphics files
  \DeclareGraphicsExtensions{.pdf,.png,.jpg,.jpeg} % for pdflatex we expect .pdf, .png, or .jpg files
\else%                                 % else we use pure latex
  \usepackage{graphicx}                % allow us to embed graphics files
  \DeclareGraphicsExtensions{.eps}     % for pure latex we expect eps files
\fi%

\graphicspath{{figures/}{pictures/}{images/}{./}} % where to search for the images

\usepackage{microtype}                 % use micro-typography (slightly more compact, better to read)
\PassOptionsToPackage{warn}{textcomp}  % to address font issues with \textrightarrow
\usepackage{textcomp}                  % use better special symbols
\usepackage{mathptmx}                  % use matching math font
\usepackage{times}                     % we use Times as the main font
\usepackage{cite}                      % needed to automatically sort the references
\usepackage{tabu}                      % only used for the table example
\usepackage{booktabs}                  % only used for the table example
\usepackage[ruled,vlined,linesnumbered]{algorithm2e}
\usepackage{subfigure}
\usepackage{amsmath}
\usepackage{amssymb}
\usepackage[inline]{enumitem}
\onlineid{6171}

\vgtccategory{Research}

\vgtcinsertpkg

\newcommand{\sysname}{KS-APR\xspace}
\title{KS-APR: Absolute Pose Regression Pipeline Keyframe Selection for Markerless Mobile AR}

\author{Changkun Liu\thanks{e-mail: cliudg@connect.ust.hk}\\ %
        \scriptsize HKUST %
\and Yukun Zhao\thanks{e-mail: yzhaoeg@connect.ust.hk}\\ %
     \scriptsize HKUST %
\and Tristan Braud\thanks{e-mail: braudt@ust.hk}\\ %
     \scriptsize HKUST}

\abstract{Markerless Mobile Augmented Reality (AR) aims to anchor digital content in the physical world without using specific 2D or 3D objects. Absolute Pose Regressors (APR) are end-to-end machine learning solutions that infer the device's pose from a single monocular image. Thanks to their low computation cost, they can be directly executed on the constrained hardware of mobile AR devices. 
However, APR methods tend to yield significant inaccuracies for input images that are too distant from the training set. 
This paper introduces KS-APR, a pipeline that assesses the reliability of an estimated pose with minimal overhead by combining the inference results of the APR and the prior images in the training set.
Mobile AR systems tend to rely upon visual-inertial odometry to track the relative pose of the device during the experience. 
As such, KS-APR favours reliability over frequency, discarding the unreliable poses.
This pipeline can integrate most existing APR methods to improve accuracy by filtering unreliable images with their pose estimates. We implement the pipeline on three types of APR models on indoor and outdoor datasets. The median error on position and orientation is reduced for all models and the proportion of large errors is minimized across datasets. Our method enables state-of-the-art APRs such as {DFNet}$_{dm}$ to outperform single-image and sequential APR methods. These results demonstrate the scalability and effectiveness of KS-APR for visual localization tasks that do not require one-shot decisions.

} % end of abstract

\CCScatlist{
  \CCScatTwelve{Visual Localization}{Camera pose regression}{Keyframe selection}{Visual positioning system};
  \CCScatTwelve{Computer Vision}{Augmented Reality}{}{}
}

\begin{document}


\firstsection{Introduction}

\maketitle

Anchoring augmented reality (AR) content without markers is an essential problem for enabling persistent and shared AR experiences. To achieve this goal, it is necessary to estimate the device's six degrees of freedom (6DoF) position and orientation in a world coordinate system, also known as absolute camera pose. In the case of mobile AR, absolute camera pose estimation faces significant constraints. The constrained hardware of mobile AR devices requires either offloading the pose estimation task to remote servers or relying on techniques with a low storage, computation, and energy footprint. A high pose estimation accuracy is also critical to avoid alignment problems between the digital content and the user's view of the physical world. Finally, mobile AR applications may involve significant mobility, requiring frequent recalibration of the local tracking mechanisms~\cite{yu2022improving, bao2022robust}, and thus low end-to-end latency.

%The absolute camera pose describes a camera's six degrees of freedom (6DoF) position and orientation in a world coordinate system. 
%Absolute camera pose estimation is an essential problem for anchoring augmented reality content without placing dedicated markers in the environment.
%Estimating the pose of a camera with a single frame is a core component in many applications, including augmented reality, navigation, and mobile robotics.
\begin{figure*}[!h]
  \centering
  \includegraphics[width=.9\linewidth]{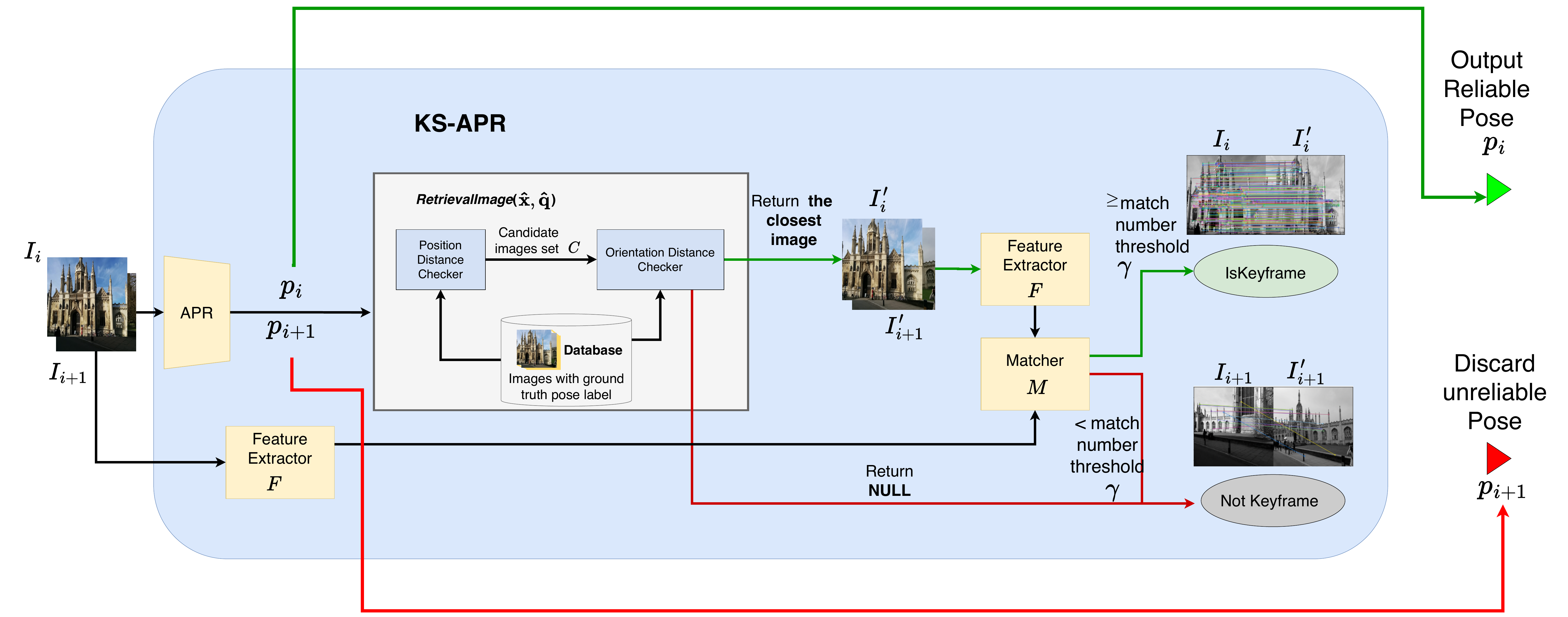}
  \caption{Keyframe selection pipeline for APR with feature matching filter. The database stores the training set images and their corresponding ground truth poses. The green and red flows represent independent branches, depending on whether an image is a keyframe. Query image $I_{i}$ is a keyframe but query image $I_{i+1}$ is not.}
\label{fig:retr}
\end{figure*}

There are currently two primary approaches to absolute pose estimation. 
Structure-based localization methods have achieved great accuracy in pose estimation. Specifically, hierarchical localization (Hloc)  pipelines~\cite{dusmanu2019d2,sarlin2019coarse,taira2018inloc,noh2017large} %representatives of this type of approach. These approaches 
implement image retrieval (IR) to fetch images similar to the query image, followed by local feature extraction and matching to establish the 2D-3D correspondences with a 3D point cloud model. %Then 2D-2D  matches are used to estimate the camera pose with Perspective-n-Point (PnP) and RANSAC based on 2D-3D correspondences via 3D point cloud models.
However, such pipelines are often computationally intensive and take hundreds of milliseconds to provide results. %This means that there is still a gap with real-time visual localization tasks on mobile devices. 
Absolute pose regressors (APR) are learning-based methods that regress the camera pose with a single monocular image input. APR methods make decisions at least one order of magnitude faster than Hloc, making them ideal for mobile and embedded applications. However, they offer lower median accuracy and are more prone to inferences with larger errors of more than 5 meters and 10 degrees. 
APR methods thus tend to lack generalisability beyond their training data, while Hloc methods provide higher accuracy and robustness~\cite{sattler2019understanding,ng2021reassessing}.
To alleviate the shortcomings of APR, methods such as augmenting the training set with labelled synthetic images~\cite{naseer2017deep,wu2017delving,sattler2019understanding}, incorporating novel view synthesis~\cite{ng2021reassessing,chen2022dfnet}, or finetuning the APR with unlabeled images~\cite{brahmbhatt2018geometry,chen2021direct,chen2022dfnet} have been proposed.
However, these techniques either introduce longer computation times or do not result in a satisfying increase in performance. Overall, neither structure-based nor APR methods can address the specific requirements of mobile AR.

% IR methods are 
% applied for exploring the limitation of APR methods because camera pose can be estimated by taking the pose
% of the most similar fetched image, or by interpolating the
% poses of several visually close images.
% Satter \emph{et al.}~\cite{sattler2019understanding}  designed an IR-based baseline for camera pose regression and found that no regressor was able to consistently surpass it on multiple localization tasks at that time.  
% This experiment shows that there is no guarantee that APR methods generalize beyond their training data.

%In this paper, we argue that APR methods' fast inference speed and low storage overhead outweigh their constraints for real-time visual positioning on mobile devices such as mobile AR. 
Most absolute pose estimation methods aim to achieve high accuracy on all input images. However, the typical modern AR pipeline involves some form of visual-inertial odometry (VIO) to track the local camera pose across the experience. As such, absolute pose estimation is only required to initially align the local experience reference system and the world reference system and periodically recalibrate the experience if needed. We thus argue that identifying accurate pose estimations and discarding those deemed unreliable while the VIO tracks the local pose is critical to ensure a good experience~\cite{yu2022improving, bao2022robust}. 

This paper introduces \sysname, a framework to estimate the accuracy of pose estimations provided by APR methods. APR methods present fast inference speed and low storage overhead, making them ideal candidates for real-time visual positioning on mobile devices such as mobile AR. However, they tend to overfit their training set, leading to significant inaccuracy with images that differ significantly from the training images, and thus an overall lower accuracy than structure-based methods.
\sysname thus identifies images close to the training set to trigger absolute pose estimation and rely on local VIO tracking between these images. \sysname relies on  the predicted 6-DoF pose of the query image to identify the closest images in the training set and evaluates their similarity through feature-based methods. Input frames with low similarity with the closest training set images are discarded. 
Hierarchical pipelines first perform image retrieval to identify the closest image in the training set, which is then used to estimate the input image's pose. \sysname performs both steps as one, leveraging APR methods' fast pose estimation time to identify the closest image in the training set. This is then used to assess the pose estimation accuracy.
As such, it is significantly faster than hierarchical pipelines while drastically improving the accuracy and robustness of existing APR methods. Identifying how close the query image is to the training set also addresses the issue of environments that evolve over time. A query image that is geographically close but significantly different feature-wise will be considered unreliable and thus discarded in favor of another query image similar to the training set.
Besides, \sysname is APR-agnostic, which allows reinforcing most existing algorithms with minimal computation and storage overhead.

We summarize our main contributions as follows:
\begin{enumerate}
    \item  We introduce a new method to determine the similarity between query images and images in a training set. We design a featureless IR algorithm using the APR 6DoF result together with this method.  
    \item   We design an APR-agnostic pipeline that implements the IR algorithm to fetch similar images in the training set, followed by local feature extraction and matching to identify reliable pose estimations. 
    \item We implement and evaluate our pipeline over three APR models on indoor and outdoor datasets. %Our proposed method outperforms both ARP and IR-based methods in terms of accuracy and robustness. 
    Our method only adds 15 ms latency to the APR runtime and improves the accuracy of original SOTA single-image APR DFNet$_{dm}$ by reducing as much as 28.6\% on position error and 22\% on orientation error in 7Scenes dataset~\cite{glocker2013real,shotton2013scene} while eliminating large errors on some scenes. %The almost negligible extra runtime less than 15$ms$ also proves that our pipeline's practical usefulness.

\end{enumerate}
\maketitle

%% \section{Introduction} %for journal use above \firstsection{..} instead
%This template is for papers of VGTC-sponsored conferences which are \emph{\textbf{not}} published in a special issue of TVCG.

%%%%%%%%% BODY TEXT
\section{Related work}

There is an extensive body of work on absolute pose estimation. This section focuses on end-to-end approaches, starting with absolute pose regression. We then review the most prominent works on image retrieval before discussing the recent studies aiming to assert the reliability of APR pose estimations.

%This section reviews the absolute pose regression, image retrieval methods, and Uncertainty estimation related to our work before motivating our keyframe selection approach.

\subsection{Absolute Pose Regression}
\label{subsec:apr}
APRs train Convolutional Neural Networks (CNN) for regression to predict the 6-DOF camera pose of a query image. The seminal work in this area is introduced by PoseNet~\cite{kendall2015posenet}. Further modifications to PoseNet have been made either by varying in architectures of network backbone~\cite{walch2017image,melekhov2017image,wu2017delving,shavit2021learning}, or by proposing different strategies in training~\cite{kendall2016modelling,kendall2017geometric}.~\cite{brahmbhatt2018geometry,radwan2018vlocnet++,valada2018deep} add visual odometry but suffer from long-term drifting in global translation and orientation. MS-Transformer (MS-T)~\cite{shavit2021learning} extends the single-scene paradigm of APR for learning multiple scenes in parallel. Rather than use a single image,~\cite{brahmbhatt2018geometry,clark2017vidloc,radwan2018vlocnet++,valada2018deep} localise sequences of images. However, all supervised APR may result in overfitting the training data, where the prediction of test images cannot extrapolate well beyond the training set~\cite{sattler2019understanding}.
To address this issue,~\cite{naseer2017deep,wu2017delving,sattler2019understanding,moreau2022lens} increase the training data size with synthetic images. Ng \emph{et al.},~\cite{ng2021reassessing} generate new camera poses to address the mismatch between the training and test distributions. MapNet+~\cite{brahmbhatt2018geometry} and MapNet+PGO~\cite{brahmbhatt2018geometry} train APR on unlabeled
video sequences. Direct-PoseNet (Direct-PN)~\cite{chen2021direct} adapts photometric loss by applying unlabeled data with NeRF synthesis in a semi-supervised manner. $\text{DFNet}$~\cite{chen2022dfnet} uses a histogram-assisted NeRF as a view renderer.  $\text{DFNet}_{dm}$~\cite{chen2022dfnet} is trained with a direct feature matching scheme to learn on unlabeled data. It achieves state-of-the-art (SOTA) accuracy by outperforming existing single-image APR methods by as much as 56\%.

Augmenting the dataset with synthetic image rendering from NeRF minimizes the overfitting of APR methods. However, it dramatically increases the model training cost. Besides, although these methods present high median accuracy performance, some estimated poses still present large errors.
% we can see that in Table~\ref{tab:dfdm_7s_level} and Table~\ref{tab:transdm_cam_level}, even for methods with very high median accuracy performance, there are still some inferences with large errors.
This paper improves the APR results' reliability by filtering out images and their pose estimates with a low probability of being accurate with low extra overhead.

% The similarity measurement method is described in detail in Section~\ref{sec:method}. We aim to pick the keyframe to keep the images with small errors and filter out the images with large errors. An image that is rich in feature points, but have never appeared in the training set is difficult for a learning-based ARP method to get the exact pose.

%Therefore, the keyframe selection in the proposed ARP method is completely different from SLAM in terms of purpose and  measurement criteria. %Keyframe selection for ARP requires new designs on algorithms. 

\subsection{Image Retrieval}
Image Retrieval (IR) is commonly used for place recognition and visual localization. Place recognition determines which part of the scene is visible~\cite{sattler2016large,weyand2016planet}, and visual localization approximates the pose of the test image by the most similar fetched images using global image descriptors. In~\cite{camposeco2018hybrid,zheng2015structure}, feature matches are used between the query image and the retrieved database image for more precise results. For visual localization tasks, IR has been employed in both structure-based hierarchical localization approaches~\cite{dusmanu2019d2,sarlin2019coarse,taira2018inloc,noh2017large} and relative pose regression (RPR) models~\cite{balntas2018relocnet,ding2019camnet,laskar2017camera,cai2019hybrid}. 
Sattler \emph{et al.}~\cite{sattler2019understanding} points out the limitations of APR compared to an IR-based baseline. APR is unable to establish a rigorous relationship between features and 3D geometry. Instead, APR is similar to image retrieval methods to estimate the pose based on the position of similar images in training set.  Despite these limitations, APRs have the upper hand over 3D structure-based methods regarding storage, inference speed, and computational complexity.
%Sattler \emph{et al.}~\cite{sattler2019understanding} shows that CNN-based APR methods are closer to IR than 3D geometry, and compares the IR-based baseline with APR, which  shows that APR methods fail to consistently outperform IR-based approaches.
%illustrates the limitation of APR as no regressor outperforms IR-based approaches consistently. 
%\tristan{Although structure-based methods often outperform APRs, they tend to be computation intensive and require storing a large 3D point cloud in addition to the reference image database.
%With the inspiration of this relation, 
Ding \emph{et al.}~\cite{ding2019camnet} thinks image retrieval
model and pose regression model should focus on different features since the former should focus more on the similarity of scenes rather than the small differences in the pose. In contrast, the latter focuses more on the small differences in pose between images. Therefore,
unlike using shared visual features for image retrieval, we develop an intermediate approach relying on pose-based IR using the result from the APR to identify reliable pose estimations.
Our pose-based IR is feature-less as shown in Algorithm~\ref{alg:retrieval}. This IR method is much faster than the visual feature-based IR method.

\subsection{Keyframe Selection and Uncertainty estimation}
Keyframe selection methods have been widely adopted in Simultaneous Localization and Mapping (SLAM)~\cite{tan2013robust,mur2015orb,mur2017orb}. These solutions often focus on unique feature points for relocalization and loop closing. %providing feature points for relocalisation. %In addition, newly selected keyframes need to be considered in connection with historical keyframes in SLAM.
However, most APR methods are based on CNN and are memoryless. A
feature-rich image that never appeared in the training set challenges learning-based APRs.
% image that is rich in feature points, but have never appeared in the training set is difficult for a learning-based APR method to get the exact pose. 
   Sattler \emph{et al.}~\cite{sattler2019understanding} emphasize the lack of guarantees for APR methods to generalize beyond their training data. Therefore keyframe selection is an entirely different process for APR methods.

In APR, keyframe selection can be achieved with uncertainty estimation.  Bayesian PoseNet~\cite{kendall2016modelling} models the uncertainty by measuring the variance of several inferences of the same input data through Monte Carlo Dropout.  Similarly, AD-PoseNet~\cite{huang2019prior} evaluates pose distribution by generating multiple poses via prior guided dropout. CoordiNet~\cite{moreau2022coordinet}  learns heteroscedastic uncertainty as an auxiliary task during the training. Poses and uncertainties output by CoordiNet are fused into an Extended Kalman Filter (EKF) to smooth the trajectories.  Deng \emph{et al.}~\cite{deng2022deep,bui20206d} represents uncertainty by predicting a mixture of multiple unimodal distributions.  Zangeneh \emph{et al.}~\cite{zangeneh2023probabilistic} follows a similar idea using a variational approach to produce arbitrarily shaped pose distributions. Although these APR methods provide both predictions of pose and estimates of uncertainty, the accuracy of estimate poses is much lower than other popular APR methods which only output poses, as shown in Section~\ref{sec:exp}.

Other works validate each estimated pose in test time with the help of Novel View Synthesis methods. 
Zhang \emph{et al.}\cite{zhang2021reference}, Taira \emph{et al.}\cite{taira2018inloc}, and Chen \emph{et al.}~\cite{chen2023refinement} render a synthesized 3D model view to verify the estimated pose of the query image and iteratively minimize the gap between the query image and the synthetic image at test time. However, the process of volume rendering and the iterative optimization algorithm is very time-consuming~\cite{moreau2022lens,chen2022dfnet}, and high storage requirements make APR even less cost-effective than traditional methods. These methods are not feasible in mobile AR applications.

Instead of modeling uncertainty during training or verifying estimated poses via novel view synthesis tools, we define the images with greater similarity to the training set as keyframes. The APR only performs pose regression on keyframes, which are the most likely to lead to small errors and filter out images that result in large errors. Our keyframe selection method identifies images that are not close to the training set, minimizing large errors and improving the overall accuracy with a low computation and storage overhead. Besides, our pipeline is APR-agnostic and very flexible; both uncertainty-aware and uncertainty-unaware pre-trained APR models can be conveniently integrated in our pipeline without relying on specific network architecture or loss functions.
\section{Proposed Approach}
\label{sec:method}

%Due to their the limited generalization capability, 
Most learning-based APR methods lack generalizability and tend to overfit their training set. %\lck{At the same time, 
Due to the distribution disparity between the training set and test set~\cite{ng2021reassessing}, these methods produce large error predictions with some query images. 
We aim to enhance the robustness of APR methods with low additional overhead by identifying whether the estimated pose of an image is reliable. 
We define as a keyframe any input image whose pose is considered reliable by our keyframe selection pipeline shown in Figure~\ref{fig:retr}. To establish whether an image is a keyframe, the proposed method finds the closest image in a database composed of the training set images and their associated ground truth pose and calculates its similarity with the query image (see Equation \ref{eq:simi}).
%We  evaluate the similarity between a query image and the closest image in the training set by equation (\ref{eq:simi}). 
%Equation \ref{eq:simi} evaluates the similarity between a query image and the closest image in the database by considering their distance in terms of position and orientation, and the number of common visual features. 
If the similarity is high, we consider the estimated pose of the query image to be reliable and identify it as a keyframe. Otherwise, the estimated pose is discarded.
%The features of both images are then matched to assess the predicted pose's accuracy. If the number of matched keypoints is large enough, the image is selected as a keyframe. 
 %Notice that for query images that cannot the retrieved image, they are directly identified as not a keyframe.

The keyframe selection pipeline operates as follows:
\begin{enumerate}
\item Given a query image $I$, APR $P$ outputs estimated position $\mathbf{\hat{x}}$ and rotation $\mathbf{\hat{q}}$ so that $P(I) = <\mathbf{\hat{x}},\mathbf{\hat{q}}>$.  
%   \item For a query image $I$, the output of APR $P(I)$ is the predicted pose $\mathbf{\hat{x}}$ and $\mathbf{\hat{q}}$ from which $I$ is taken.
    \item A feature extractor $F$ extracts the features keypoints and descriptors of image $I$  as $F(I)$.
    If an image $I'$ is retrieved successfully from the database by Algorithm \ref{alg:retrieval}, \emph{RetrievalImage,}($\mathbf{\hat{x}},\mathbf{\hat{q}}$), proceed to step 3. Otherwise, \emph{RetrievalImage}($\mathbf{\hat{x}},\mathbf{\hat{q}}$) returns NULL since none of the images in the database meets the distance requirement, and $I$ is identified as not a keyframe. % directly. Jump to 1. for the next query image in this case.
%    Run \emph{RetrievalImage} ($\mathbf{\hat{x}},\mathbf{\hat{q}}$) in database, the result $I'$ is the most similiar valid image or an invalid image NULL.
    \item The feature extractor $F$ extracts the feature on $I'$ and gets the keypoints and associated descriptors as $F(I')$. A matcher $M$ matches the features on $F(I)$ and $F(I')$. We note good matches as  $M(F(I),F(I'))$.
    If the number of good matches %\lck{In OpenCV websites, it use good matches} 
    $|M(F(I),F(I'))|$ is larger than the match number threshold $\gamma$, the pipeline identifies $I$ as a keyframe, otherwise $I$ is not a keyframe and is discarded.
\end{enumerate}

Figure~\ref{fig:retr} illustrates both the case where a query image $I_{i}$ is identified as a keyframe with a retrieved image $I'_{i}$ and the case where another query image $I_{i+1}$ is considered not to be a keyframe with a retrieved image $I'_{i+1}$.

Compared to other state-of-the-art structure-based localization approaches such as Hloc pipelilnes~\cite{sarlin2019coarse}, our method is APR-agnostic and relies on feature-less retrieval of closest images. Algorithm~\ref{alg:retrieval} displays the detailed operation of the image retrieval process. This algorithm only requires the 6-DoF camera pose $<\mathbf{\hat{x}}$, 
$\mathbf{\hat{q}}>$ of the image estimated by any APR methods. The search for a candidate image is based on the distance between the predicted pose $<\mathbf{\hat{x}}$, 
$\mathbf{\hat{q}}>$, and the training set's image poses. By relying on 6-DoF poses rather than high-dimensional visual feature descriptors, our proposed method significantly decreases the size of the search space.

\begin{algorithm}[t]
  \SetAlgoLined
  \KwData{Query image $I$, APR Output $P(I)$ = $<\mathbf{\hat{x}},\mathbf{\hat{q}}>$.}
  Initialize an empty closest candidate images set $C$\;
Initialize the minimal orientation distance $\Delta_{\mathbf{\hat{q}}{\text{min}}}\leftarrow \infty$\;
// $i$ is the index of images in database\;
  \For{$i = 0,1,2,\cdots$}
  {
     \If{$||\mathbf{\hat{x}}-\mathbf{x}_i||_2 \leq d_{th}$,}
     {Add $\mathbf{q}_i$ of image $I_i$ in $C$\;}
   }
    \uIf{$C \text{ is } \emptyset$}
    {Return \textbf{NULL}\;}
    \Else
    {
    \For{$\mathbf{q}_j$ \text{ in } $C$}
    {\If{$||\frac{\mathbf{\hat{q}}}{||\mathbf{\hat{q}}||}-\mathbf{q}_j||_2 \leq \Delta_{\mathbf{\hat{q}}{\text{min}}}$}
       {$\Delta_{\mathbf{\hat{q}}{\text{min}}} \leftarrow ||\frac{\mathbf{\hat{q}}}{||\mathbf{\hat{q}}||}-\mathbf{q}_j||_2$\;
       $I_{closest} \leftarrow I_j$\;}
    }
    {Return $I_{closest}$}
    }
\caption{RetrievalImage($\mathbf{\hat{x}},\mathbf{\hat{q}}$)}
\label{alg:retrieval}
\end{algorithm}

%The most important step in this pipeline is step.2, where we use the feature-less retrieval image Algorithm. \ref{alg:retrieval}. 
%We design this algorithm that only relies on the 6-DoF camera pose of the image estimated by any APR methods. 
%A simple search by matching the query image with the database images based on $\mathbf{\hat{x}}$ and  
%$\mathbf{\hat{q}}$ is performed. The search based on 6-DoF camera pose is very efficient compared to the way using feature descriptors.

\subsection{Image Retrieval based on APR}
\label{subsec:imgretrieval}
% We explain our purpose of designing this pipeline and the details of the Algorithm. \ref{alg:retrieval} in this section.  

Let $\Delta_{\mathbf{\hat{x}}}$ and $\Delta_\mathbf{\hat{q}}$ be the Euclidian distance between the pose of image $I$ pose with a ground truth pose $< \mathbf{x'}, \mathbf{q'} >$ of an image $I'$ in database:
\begin{align}
\Delta_{\mathbf{\hat{x}}} = ||\mathbf{\hat{x}}-\mathbf{x'}||_2 \\
\Delta_\mathbf{\hat{q}} = ||\frac{\mathbf{\hat{q}}}{||\mathbf{\hat{q}}||}-\mathbf{q'}||_2
\label{eq:deltaxq}
\end{align}

\begin{figure}[!t]
  \centering
  \includegraphics[width=.9\linewidth]{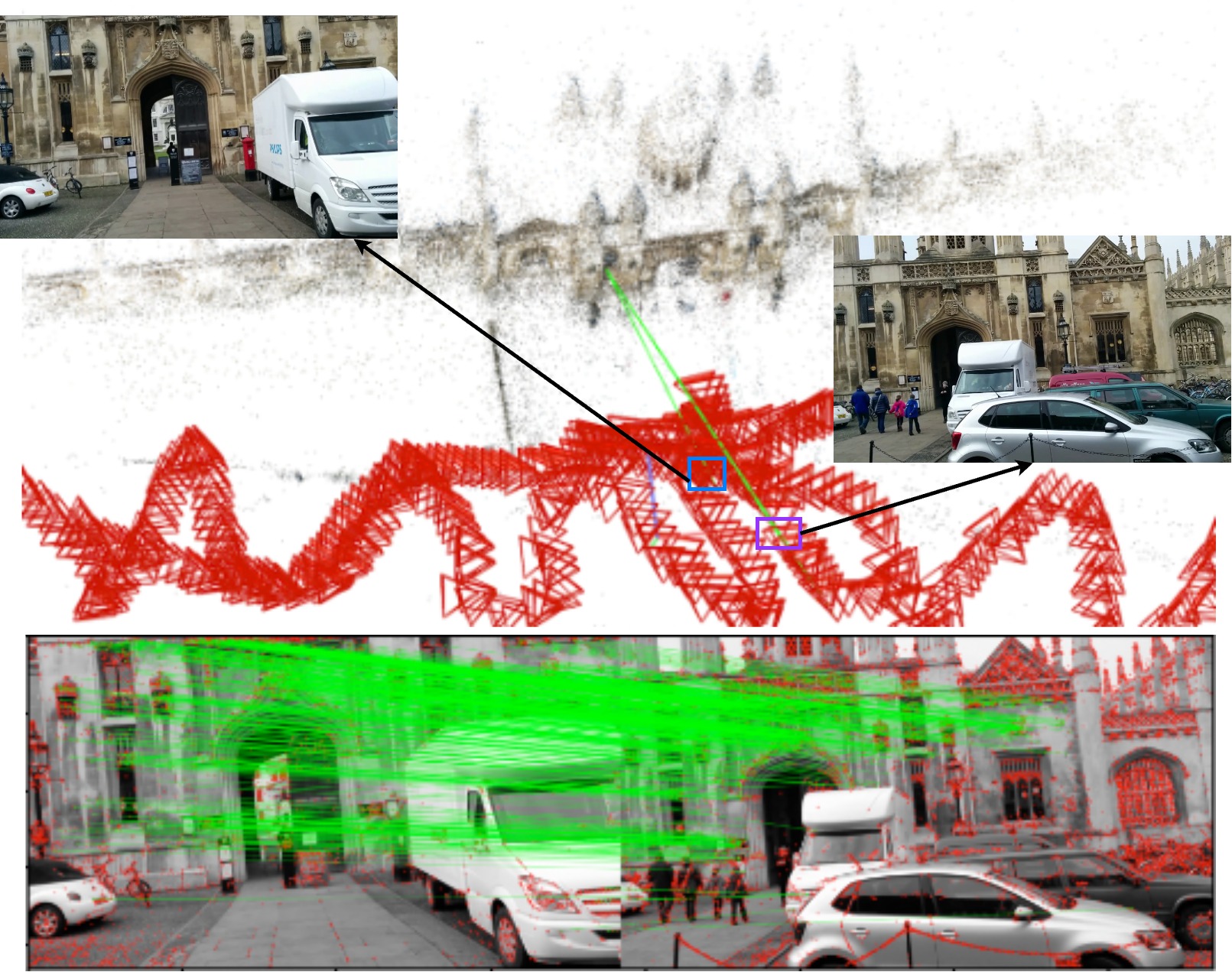}
  \caption{COLMAP~\cite{schonberger2016structure} %for Structure From Motion (SFM) 
  3D reconstruction of the images in KingsCollege from Cambridge dataset~\cite{kendall2015posenet}. % some of the same feature points come from images that are far apart, as shown in the blue and purple boxes. 
  Two images presenting many feature matches may not be close in spatial location (e,g, blue and purple box), leading to high inaccuracy in most APR.
  %Even though these two images contain many common feature points, it does not mean that these two images are close in spatial location. 
  }
\label{fig:methodex}
\end{figure}

%\lck{Given the limitations of learning-based APR~\cite{ng2021reassessing},  our pipeline identifies keyframes by searching images in database based on their distance to the query image as a preliminary step before further feature extraction and feature matching. It filters unreliable images both in terms of pose and number of feature matches with fast execution.} %in addition to the number of matching features results in faster search and better accuracy than solely the number of feature matches.

%We argue that searching images based on their distance to the query image, followed by assessing the number of matching features results in faster search and better accuracy than solely the number of feature matches. %\lck{This sentence alters our original meaning.}
%\lck{As such, our search for the most similar images is based on the estimated position and orientation of query images output by APR. }

Let us consider the example presented in Figure~\ref{fig:methodex}. The two selected images have a high number of feature matches. However, they are separated by 6.8 meters and 4.7 degrees in position and rotation, respectively. Ng \emph{et al.}~\cite{ng2021reassessing} have shown that existing learning-based methods fail to localize images with a large deviation in position and orientation compared to the images in the training set.
Therefore, our pipeline focuses on finding highly similar images in terms of pose and number of feature matches.
% As shown in Figure~\ref{fig:methodex}, these two camera poses have an error of 6.8 meters and 4.7 degrees in position and orientation, respectively, but they have many common features. \cite{ng2021reassessing} already proved that existing learning-based methods fail to learn or localize images with a large bias in the position and orientation of the images in the train set. 
We express the degree of similarity between $I$ and $I'$, $S(I,I')$ as a sequential three-parts measure in this paper: (1) their distance in terms of position %\lck{'both' is a bit weird?}
(2) their distance in terms of orientation, and (3) the number of feature matches.
\begin{align}
\begin{split}
S(I,I')= \left \{
\begin{array}{ll}
    0,    & \Delta_{\mathbf{\hat{x}}}> d_{th}\\
    0, &\mathbf{q'} \neq \arg\min\limits_{\Delta_{\mathbf{\hat{x}}}\leq d_{th}}(\Delta_\mathbf{\hat{q}})\\
    |M(F(I),F(I'))|, & \mathbf{q'} = \arg\min\limits_{\Delta_{\mathbf{\hat{x}}}\leq d_{th}}(\Delta_\mathbf{\hat{q}})
\end{array}
\right.
\end{split}
\label{eq:simi}
\end{align}
where $d_{th}$ is the threshold of position distance in Algorithm~\ref{alg:retrieval}. The similarity $S(I,I')$ of any image $I'$ in the database to a  query image $I$ is bound by  % equation (1), equation (2) and 
equation (3). %The similarity between any image $I'$ in the database and $I$ can only be within these three conditions, as shown in equation (3). 
The algorithm first checks whether $\mathbf{x'}$ of $I'$ is close enough from $\mathbf{\hat{x}}$ of $I$. If this condition is not met, the similarity function returns 0; otherwise, the function checks whether  $\mathbf{q'}$ is close enough to $\mathbf{\hat{q}}$. If this second condition is not met, the similarity function returns 0; otherwise, the function returns the number of good feature matches between $I$ and $I'$. 
%An image $I'$ resulting in $S(I,I') = 0$ will not be fetched by Algorithm \ref{alg:retrieval}.

% $\mathbf{x'}$ of $I'$ does not meet condition 1 in equation (3), is 0. Secondly, $\mathbf{x'}$ meets the condition 1 in equation (3) while $\mathbf{q'}$ is not the closest $\mathbf{q'}$ to $\mathbf{\hat{q}}$ in images meets condition 1, that is condition 2, so is 0. Thirdly, $\mathbf{x'}$ meets the condition 1, and $\mathbf{q'}$ is the closest $\mathbf{q'}$ to $\mathbf{\hat{q}}$ in images meets condition 1, that is condition 3, then $S(I,I')$ is the number of good matches. $I'$ has $S(I,I') = 0$ will not be fetched by Algorithm \ref{alg:retrieval}.}
% For example, if $S(I,I') \geq \gamma$ with $\gamma$ the match number threshold in the third stage  of our pipeline, then $I$ is identified as a keyframe. Otherwise it is filtered out. Compared with traditional feature match localization methods, the image retrieval algorithm we propose does not focus on obtaining images sharing enough features. It rather focuses on detecting highly similar images in the training set which   

% What we want to detect is whether we can find an image in the training set that is very similar to the query image based on equation. \ref{eq:simi}.

Accordingly, Algorithm~\ref{alg:retrieval}  retrieves the image by the estimated pose based on Equations (1), (2), and (3). An image $I'$ resulting in $S(I,I') = 0$ will not be fetched by Algorithm \ref{alg:retrieval}.
From line 4 to line 8, the algorithm first checks whether the training set contains images near the estimated location (condition 1 of Equation (3)). Then it selects all the images in the training set that are close to this position and selects the one with the closest orientation in lines 11 to 19 (condition 2 of Equation (3)).  
As such, the algorithm either returns the closest image in terms of position and orientation or NULL if no close image has been identified. 
When the algorithm returns NULL, the query image is thus either too far from the training set to be relevant or the predicted pose is not within the average accuracy of the APR method.

\subsection{Feature matching}

If a database image $I'$ is close enough to the query image $I$ in terms of position and rotation, the pipeline matches the features between $I$ and $I'$ as $|M(F(I),F(I'))|$. The query image is considered a keyframe if:

\begin{equation}
    |M(F(I),F(I'))|\geq\gamma, \gamma \in \mathbb{N}^+
\end{equation}
where $\gamma$ is a threshold that depends on the average feature richness (number and uniqueness) of the training set. 

\subsection{Outcomes}

%Now we perform an intuitive verification of the algorithm and the pipeline. 
The proposed keyframe selection pipeline may filter out a query image $I$ under two scenarios:
\begin{enumerate*}
    \item The training set does not contain an image close enough to the estimated location $\mathbf{\hat{x}}$.
    \item Although a similar image is in the database based on $\mathbf{\hat{x}}$ and $\mathbf{\hat{q}}$, it presents too few good feature matches.
\end{enumerate*}
In the first scenario, $I$ cannot be recognized as a keyframe because of the limited generalization ability of the neural network. 
The APR thus has a high chance of predicting a vastly incorrect pose for an image $I$ far from the training set.
%and failure to find the image means that $I$ is far from the training set.
%and the probability that $<\mathbf{\hat{x}}, \mathbf{\hat{q}}>$ has a large error will be high. 
In the second scenario, although a valid most-similar image $I'$ is found in the training set based on $<\mathbf{\hat{x}}, \mathbf{\hat{q}}>$, it does not present enough feature matches. Either the orientations of the two images are very different, leading to little overlap, or the predicted  $\mathbf{\hat{x}}$ and $\mathbf{\hat{q}}$ have a large error, and the image $I'$ found according to this pose is a false positive. We further analyze these conditions in Section \ref{subsec:analy}.% display the capabilities of our algorithm and analyze results in detail in Section \ref{sec:exp}.

\subsection{Hyperparameters}
\label{sec:hyper}
The proposed method features two hyperparameters that should be tuned to the scene: the distance threshold $d_{th}$, and the minimum number of feature matches $\gamma$. $d_{th}$ depends on the size of the scene and the position error threshold we think is large. If $d_{th}$ is much smaller than the median position error of the APR,  query images' poses may be wrongly identified as incorrect. If $d_{th}$ is much larger than the median position error, images with large estimated errors or far away from the training set will be considered keyframes. For 7Scenes dataset (see Section~\ref{sec:exp}), $d_{th}$ ranges between $0.1-0.3m$. For Cambridge dataset, $d_{th}$ ranges between $1-2m$.  We sample image pairs  ($I$ and  $I_{far}$) where $I$ comes from one sequence in the training set for simplicity and $I_{far}$ is selected through all training set. For example, we get all the pairs based on training sequence 1 of Kings using the above process with $d_{th} = 1.5$. In this set, the maximum number of matches is $25$. Therefore, we set $\gamma$ near 25. We fine tune $\gamma$ around the maximum number of matches of these pairs for each dataset. If $\gamma$ is too large,  many query images will be discarded. If $\gamma$ is too small, images with large estimated errors may be identified as keyframes.
%%$d_{th}$ depends on the size of the scene, the distance between images in the training set, and the median accuracy of the pipeline's APR in that scene.  If $d_{th}$ is much smaller than the median position error of the APR,  query images' poses may be wrongly identified as incorrect.
%%If $d_{th}$ is much larger than the median position error, images with large estimated errors or far away from the training set will be considered keyframes.
%%$\gamma$ depends on the feature richness of the scene and the accuracy of the pipeline's APR.
%%If $\gamma$ is too large,  many query images' matches will be discarded.
%%If $\gamma$ is too small, images with large estimated errors may be identified as keyframes. As different APRs fetch different images due to different errors, the final selection of $\gamma$ is also be affected by APR.
%%Following this intuition, we fine-tune\lck{fine tune?} $d_{th}$ and $\gamma$ for each scene and APR using a grid search in a small range. For 7Scenes dataset (see Section~\ref{sec:exp}), $d_{th}$ ranges between $0.1-0.3m$, and $\gamma$ between $20-70$. For Cambridge dataset, $d_{th}$ ranges between $1-2m$, and $\gamma$ between $10-60$. 

\subsection{Application of \sysname in AR applications}
Most situated mobile markerless AR applications require placing virtual objects at specific coordinates in the world coordinate systems, or at least, in the coordinates of the Structure-from-Motion (SfM) model of the scene. 
As such, they require absolute pose estimation. %Structure-based methods require extensive storage and computing power that cannot be achieved on-device. Sending the query image to a remote server raises privacy concerns. 
With recent improvements in mobile device processors, APR methods can be run in close-to-real-time on-device.

\begin{figure}[!t]
  \centering
  \includegraphics[ width=\linewidth, height=10cm]{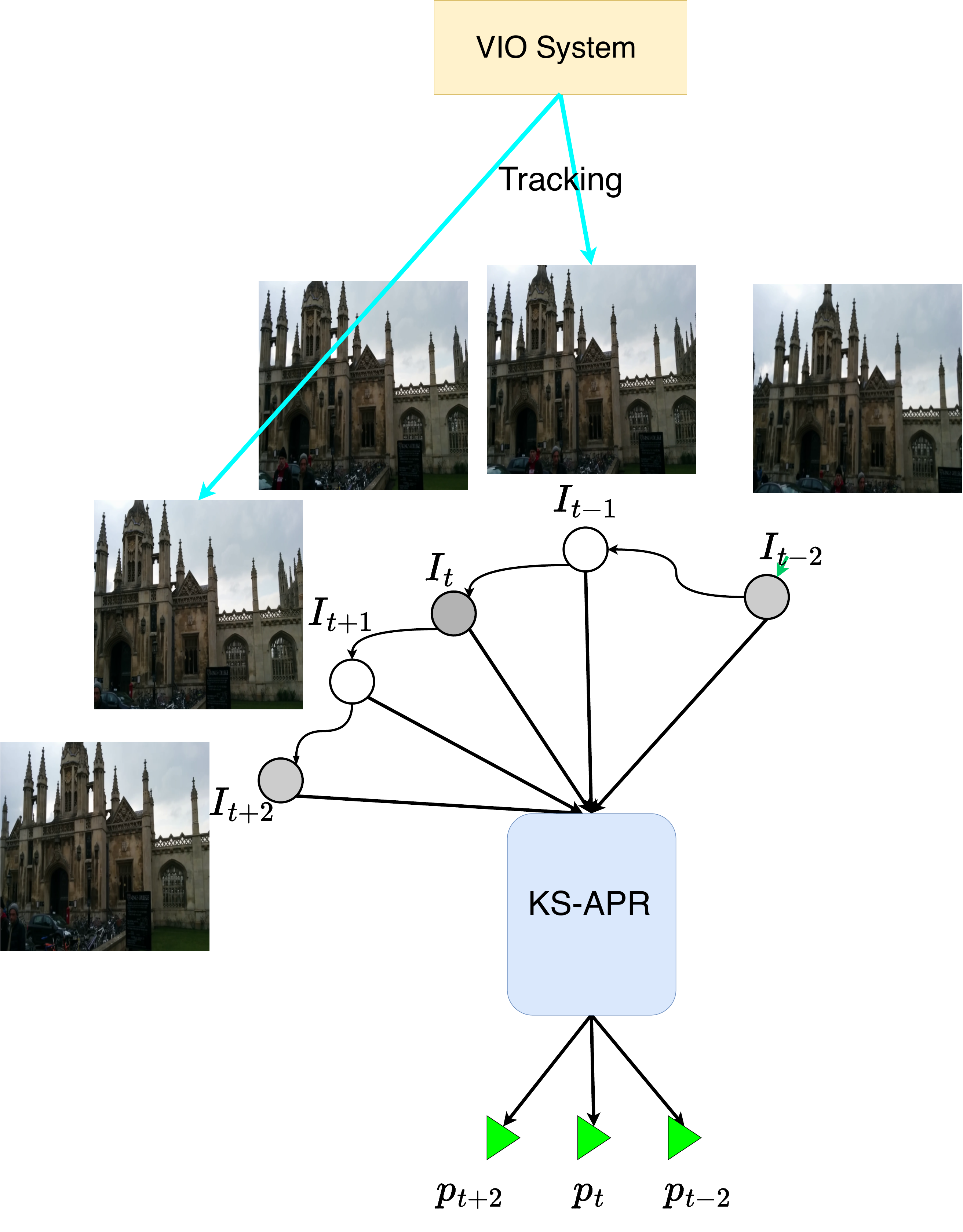}
\caption{ The framework of mobile markerless AR with \sysname. The grey circles are keyframes. White circles are not keyframes and \sysname filters out their predictions. We can use the tracking module in ARKit or ARCore to bridge keyframes and non-keyframes.}
\label{fig:framework}
\end{figure}

\sysname can thus be integrated into most current mobile AR systems such as smartphones. By discarding input frames that would lead to inaccurate results, \sysname improves the accuracy and robustness of the underlying APR, at the cost of a significant reduction in the number of input frames. However, in the case of AR applications, this operation results in compelling advantages. \sysname can be integrated with traditional VIO-based tracking such as ARkit\footnote{\url{https://developer.apple.com/documentation/arkit/configuration_objects/understanding_world_tracking}} or ARCore\footnote{\url{https://developers.google.com/ar}} to track the local pose between reliable pose estimations as shown in Figure~\ref{fig:framework}. \cite{scargill2022here} shows that ARKit can achieve small drift in centimeter-level in most cases.

Discarding unreliable query images also allows to solve the major challenge of environment changes in situated AR. Many elements change over time, in the short (pedestrians, moving vehicles, plants), medium (parked vehicles, advertisements), or long term (storefronts, buildings). With visual data being collected infrequently, this often results in a discrepancy between the training set and the physical world. \sysname discards images that are too dissimilar from the training set. Instead of providing vastly inaccurate estimation (APR) or fail registration after long computation times (structure-based), \sysname discards query images containing significant environment changes, in favor of those that primarily contain static elements.  

% %For instance, tourist AR applications 
% For AR applications that are used in tourist attractions and landmark locations, as long as the locations and orientations where people take photos most frequently are covered in the training set, \sysname can ensure output accurate poses in most cases. We can utilize tracking module in ARkit\footnote{\url{https://developer.apple.com/documentation/arkit/configuration_objects/understanding_world_tracking}} to keep the place of visual objects before obtaining a new reliable pose. 

\section{Experiment}
\label{sec:exp}
%As seen in Section~\ref{subsec:apr},  mainstream APR techniques can be divided into three main categories in terms of training methods: supervised learning APR which overfit their training sets; supervised learning APR which learn both original training sets and synthetic images, semi-supervised learning APR which learn unlabel data. 
To demonstrate the scalability and effectiveness of \sysname, we select the latest representative mainstream APR techniques and combine them with our pipeline over both small indoor datasets and large-scale outdoor datasets. 

% \subsection{Experimental Setting}
% \label{subsec:imp}
% In this section, we introduce the datasets as well as the implementation details of our keyframe selection pipeline.

%We introduce the implementation details for experiments on each data set with different APR models. %We also provide more details in the supplementary material.}

\begin{table*}[t]
\caption{\textbf{Labeled} training methods of \textbf{single-frame} APR, \textbf{IR} baseline (DenseVLAD+Inter.) and APR methods with \textbf{uncertainty estimation} (\text{Bayesian PN, CoordiNet, CoordiNet+LENS, BMDN, VaPoR}) compared with our method on the 7Scenes datasets. We report the median position/orientation errors in m/°. $\text{DFNet}^{ks}$ achieves top performance by succesfully identifying the keyframe. Best results are highlighted in bold.} % Detailed distribution of error and percentages of keyframes are detailed in Table~\ref{tab:dfdm_7s_level}. Best results are highlighted in bold.}
$$
\begin{array}{l|cccccc|l}
\hline \text{Labeled Methods \& \text{IR} } & \text{Chess } & \text{Fire } & \text{Heads } & \text{Office }&\text{Pumpkin } & \text{Stairs } & \text{Average } \\
\hline \text{PoseNet(PN)\cite{kendall2015posenet} }& 0.32/8.12& 0.47/14.4 &0.29/12.0& 0.48/7.68& 0.47/8.42& 0.47/13.8 & 0.42/10.7\\
\text{PN Learn } \sigma^2\text{\cite{kendall2017geometric}} & 0.14/4.50 &0.27/11.8 &0.18/12.1 &0.20/5.77& 0.25/4.82& 0.37/10.6 & 0.24/8.27\\
\text{geo. PN\cite{kendall2017geometric} } & 0.13/4.48 &0.27/11.3 &0.17/13.0& 0.19/5.55& 0.26/4.75  &0.35/12.4 & 0.23/8.58\\
\text{LSTM PN\cite{walch2017image} } & 0.24/5.77 &0.34/11.9 &0.21/13.7& 0.30/8.08& 0.33/7.00 &0.40/13.7& 0.30/10.0\\
\text{Hourglass PN\cite{melekhov2017image} } & 0.15/6.17& 0.27/10.8 &0.19/11.6& 0.21/8.48& 0.25/7.0 & 0.29/12.5& 0.23/9.42\\
\text{BranchNet\cite{wu2017delving} } & 0.18/5.17& 0.34/8.99 &0.20/14.2 &0.30/7.05 &0.27/5.10& 0.38/10.3& 0.28/8.47  \\
\text{MapNet\cite{brahmbhatt2018geometry} } & 0.08/3.25& 0.27/11.7 &0.18/13.3 &0.17/5.15& 0.22/4.02 &0.30/12.1& 0.20/8.25\\
\text{Direct-PN\cite{chen2021direct}} & 0.10/3.52 &  0.27/8.66  & 0.17/13.1 &  0.16/5.96 &  0.19/3.85   & 0.32/10.6  & 0.20/7.62 \\
\text{MS-T\cite{shavit2021learning}} & 0.11/4.66 & 0.24/9.60  &0.14/12.2  &0.17/5.66 & 0.18/4.44  & 0.26/8.45& -/-\\
%\text{MS-Transformer}^{ks} \text{(ours) }& 0.09/4.8(65\%) & 0.19/8.6(64\%)  & 0.12/10.6(64\%) &0.14/5.62 (60\%) & 0.13/3.8 (63\%) & 0.22/7.8(68\%) &\\
\text{DFNet\cite{chen2022dfnet} } & 0.05/1.88& 0.09/3.22& 0.06/3.63& 0.08/2.48 &0.11/2.81& 0.17/3.29& 0.09/2.89\\ \hline
\text {Bayesian PN\cite{kendall2016modelling} } & 0.37/7.24 &0.43/13.7 & 0.31/12.0 & 0.48/8.04& 0.61/7.08& 0.48/13.1 & 0.45/10.2\\
\text{CoordiNet\cite{moreau2022coordinet}} &0.14/6.7&0.27/11.6&0.13/13.6&0.21/8.6&0.25/7.2&0.28/12.9&0.21/10.1 \\
\text{CoordiNet+LENS\cite{moreau2022lens}}&\mathbf{0.03}/\mathbf{1.3}&0.10/3.7&0.07/5.8&\mathbf{0.07}/\mathbf{1.9}&\mathbf{0.08}/2.2&0.14/3.6&0.08/3.08\\
\text{BMDN\cite{bui20206d}} & 0.10/4.4 &0.28/11.9 & 0.12/12.8 & 0.19/6.6&  0.22/6.9&0.31/10.0 & 0.20/8.77\\
\text{VaPoR\cite{zangeneh2023probabilistic}} & 0.17/6.9 & 0.30/14.1&0.17/14.5 &0.24/9.3&0.30/8.3 &0.47/15.5 & 0.27/11.4 \\\hline
\text{DenseVLAD+Inter.\cite{sattler2019understanding}} & 0.18/10.0& 0.33/12.4 & 0.14/14.3& 0.25/10.1 &0.26/9.42&0.24/14.7 & 0.23/11.8\\\hline
\text{DFNet}^{ks} \text{(ours) }& 0.05/1.73& \mathbf{0.06/2.35}& \mathbf{0.05/3.14}& \mathbf{0.07}/2.22 &\mathbf{0.08/2.10}& \mathbf{0.12/2.74}&\mathbf{0.07/2.38} \\
\hline
\end{array}
$$
\label{tab:df_7s_rank}
\end{table*}

\begin{table*}[t]
\caption{\textbf{Unlabeled single-frame} APR methods  (MapNet+, MapNet+PGO, Direct-PN+U, $\text{DFNet}_{dm}$) and \textbf{sequential-based} APR (MapNet+, MapNet+PGO, VLocNet) compared with our method on 7Scenes dataset. We report the median position/orientation errors in m/° and $\text{DFNet}^{ks}_{dm}$ achieves top performance by only keeping keyframe. %Detailed distribution of error and percentages of keyframes are detailed in Table~\ref{tab:dfdm_7s_level}.
}
$$
\begin{array}{l|cccccc|l}
\hline \text{Methods } & \text{Chess } & \text{Fire } & \text{Heads } & \text{Office }&\text{Pumpkin } & \text{Stairs } & \text{Average } \\\hline
\text{MapNet+\cite{brahmbhatt2018geometry}}& 0.10/3.17 &0.20/9.04& 0.13/11.1& 0.18/5.38& 0.19/3.92 & 0.30/13.4 & 0.18/7.67\\
\text{MapNet+PGO\cite{brahmbhatt2018geometry}}& 0.09/3.24 &0.20/9.29 &0.12/8.45 &0.19/5.42 &0.19/3.96 &0.27/10.6 & 0.18/6.83\\
\text{VLocNet\cite{valada2018deep}}& 0.04/1.71&\mathbf{0.04}/5.34&0.05/6.65&\mathbf{0.04}/1.95&\mathbf{0.04}/2.28&0.10/6.48&\mathbf{0.05}/4.07\\
\text{Direct-PN+U\cite{chen2021direct}} &0.09/2.77& 0.16/4.87& 0.10/6.64 &0.17/5.04& 0.19/3.59 &0.24/8.52& 0.16/5.24 \\
\text{DFNet}_{d m}\text{\cite{chen2022dfnet}} & 0.04/1.48& \mathbf{0.04}/2.16& 0.03/1.82& 0.07/2.01 &0.09/2.26& 0.14/3.31& 0.07/2.17 \\ \hline
\text{DFNet}_{d m}^{ks}\text{(ours) }& \mathbf{0.03/1.25}&  \mathbf{ 0.04/1.60}& \mathbf{0.02/1.47} & 0.07/\mathbf{1.87} & 0.08/\mathbf{1.75}& \mathbf{0.08/2.24} & \mathbf{0.05/1.70}\\
\hline
\end{array}
$$
\label{tab:dm_7s_rank}
\end{table*}

\begin{table*}[t]
\caption{ 3D methods compared with our method on 7Scenes dataset. We report the median position/orientation errors in m/°. %Detailed distribution of error and percentages of keyframes are detailed in Table~\ref{tab:dfdm_7s_level}.
}
$$
\begin{array}{ll|cccccc|l}
\hline
 \text{Methods } & & \text{Chess } & \text{Fire } & \text{Heads } & \text{Office }&\text{Pumpkin } & \text{Stairs } & \text{Average } \\\hline
\text{Ours }&\text{DFNet}_{d m}^{ks}& 0.03/1.25&  0.04/1.60 & 0.02/1.47 & 0.07/1.87 & 0.08/1.75& 0.08/2.24 & 0.05/1.70\\
\hline
&\text {Inloc~\cite{taira2018inloc}} & 0.03/1.05 &0.03/1.07 &0.02/1.16 &\mathbf{0.03}/1.05 &0.05/1.55&0.09/2.47 & 0.04/1.39\\
\text{One-shot 3D}&\text {DSAC++~\cite{brachmann2018learning}} & \mathbf{0.02}/\mathbf{0.50} &\mathbf{0.02}/\mathbf{0.90} &\mathbf{0.01}/\mathbf{0.80}& \mathbf{0.03}/0.70& \mathbf{0.04}/1.10& 0.09/2.60 & 0.03/1.10 \\
&\text {SCoordNet~\cite{zhou2020kfnet}} & \mathbf{0.02}/0.63 &\mathbf{0.02}/0.91&0.02/1.26 &\mathbf{0.03}/0.73 &\mathbf{0.04}/1.09& 0.04/1.06 & 0.03/0.95\\
\hline
\text{Sequential 3D} & \text {KFNet~\cite{zhou2020kfnet}} & \mathbf{0.02}/0.65 &\mathbf{0.02}/\mathbf{0.90}&\mathbf{0.01}/0.82&\mathbf{0.03}/\mathbf{0.69} &\mathbf{0.04}/\mathbf{1.02}& \mathbf{0.03}/\mathbf{0.94} &\mathbf{0.02}/\mathbf{0.84}\\
\hline
\end{array}
$$
\label{tab:3d_7s}
\end{table*}

\begin{table*}[t]
\caption{Percentage of images with pose predicted with high (0.25m, $2^{\circ}$), medium (0.5m, $5^{\circ}$), and low (5m, $10^{\circ}$) accuracy~\cite{sattler2018benchmarking}. The value in parentheses represents the percentage of keyframes identified in the test set by our method.}
$$
\begin{array}{l|cc||cc}
\hline
\text{Dataset}   & \text{DFNet \cite{chen2022dfnet} }   & \text{DFNet}^{ks} \text{(ours) } & \text{DFNet}_{d m}\text{\cite{chen2022dfnet}} & \text{DFNet}_{d m}^{ks}\text{(ours) }\\ \hline
\text{Chess}  &54.4/91.7/\mathbf{99.7} &\mathbf{59.8/94.1/99.7}(71\%) &67.9/95.6/98.1 & \mathbf{78.8/99.0/99.8}(64\%) \\
\text{Fire}  &32.7/62.8/86.2 &\mathbf{43.8/74.4/92.2}(66\%) &46.2/88.2/99.6   & \mathbf{63.4/97.3/100}(57\%)\\
\text{Heads}   & 21.2/71.4/92.7&\mathbf{27.2/80.0/98.9}(65\%)&57.6/92.7/99.3 &\mathbf{72.2/99.0/100} (63\%) \\
\text{Office}  &38.6/84.6/95 &\mathbf{43.5/90.5/97.1}(60\%) & 49.7/91.4/99.2& \mathbf{54.7/93.3/100}(66\%) \\ 
\text{Pumpkin}   &34.2/68.4/85.3 &\mathbf{45.7/84.9/95.0}(68\%) &44.9/83.1/96.3 & \mathbf{57.0/94.6/99.9}(70\%)\\
\text{Stairs}    &19.4/62.2/92.6 & \mathbf{27.0/73.0/96.4} (72\%) &37.4/85.3/97.1 &\mathbf{40.7/94.1/99.8}(63\%)  \\\hline
\end{array}
$$
\label{tab:dfdm_7s_level}
\end{table*}

\subsection{Datasets}

 The 7Scenes dataset~\cite{glocker2013real,shotton2013scene} consists of seven small indoor scenes from $1m^3$
to $18m^3$. Each scene contains 1000 to 7000 images in the training set and 1000 to 5000 images in the test set. We use six of its seven scenes for comparative evaluation as the official project websites of DFNet and $\text{DFNet}_{dm}$ only provide pre-trained models for these six scenes. The Cambridge Landmarks~\cite{kendall2015posenet} dataset represents six large-scale outdoor scenes ranging from $900m^2$ to $5500m^2$. We use four of its six scenes for comparative evaluation. Each scene contains 231 to 1487 images in the training sets and 103 to 530 in the test sets. 

\subsection{Implementation Details}

\begin{table}[hbt!]
\caption{Influence of $d_{th}$ and $\gamma$ on accuracy for $\text{DFNet}_{dm}^{ks}$ in Chess. The first row is $\text{DFNet}_{dm}$.  We report the median position/orientation errors in $m/^{\circ}$, the percentage of images resulting in high (0.25m, $2^{\circ}$), medium (0.5m, $5^{\circ}$), and low (5m, $10^{\circ}$) accuracy~\cite{sattler2018benchmarking}, and the percentage of keyframes identified in the test set by our method. } 
$$
\begin{array}{c|c|c}
\hline
 d_{th}/\gamma        & m/ ^{\circ}   & \text{high/median/low~(keyframe ratio)}     \\ \hline
-/- & 0.04/1.48 & 67.9/95.6/98.1(-) \\ \hline
0.15/20   &\mathbf{0.03}/1.26 & 78.0/98.9/\mathbf{99.8}~(66\%)    \\
0.15/30   &  \mathbf{0.03/1.25} &  \mathbf{78.8/99/99.8 }~(64\%)   \\ 
0.20/30    & 0.04/ 1.36 &  74.0/97.2/99.7~(77\%)    \\ 
0.20/20    & 0.04/ 1.38 &  73.1/96.9/99.6~(79\%)    \\ \hline
\end{array}
\label{tab:paraset}
$$
\end{table}
\label{subsec:imp}
We implement \sysname as a set of python scripts that take the output pose from the selected APR method and apply the algorithms discussed in Section~\ref{sec:method}. 
There are many methods for feature extraction and matching~\cite{detone2018superpoint,harris1988combined,cieslewski2018data,sarlin2020superglue}. We use the SIFT~\cite{lowe2004distinctive} detector and descriptor and match the features using OpenCV's brute force feature matching. We filter bad matches by applying the ratio test proposed by D. Lowe~\cite{lowe2004distinctive}. We set Lowe's ratio test as 0.7 for getting good matches except for the Hospital dataset. The ratio test is 0.5 for the Hospital dataset as the scene contains many symmetrical and repetitive features (see Section~\ref{subsec:analy} for discussions).
To accelerate feature extraction and matching, we resize all images to $384\times384$, and then center crop them to $380\times 380$. All images are converted to grayscale before feature extraction.

%\subsubsection{APR Models}

We implement \sysname over three recent APR models:

\noindent\textbf{DFNet}. DFNet~\cite{chen2022dfnet} is the SOTA model in supervised learning APR that uses exposure-adaptive novel view synthesis. We use the official pre-trained model\footnote{\url{https://github.com/ActiveVisionLab/DFNet}} %that reproduces the results of the paper~\cite{chen2022dfnet} 
in our keyframe selection pipeline and apply it to the 7Scenes dataset.
%We set the match number threshold $\gamma$ between 20 to 40, and the distance threshold  $d_{th}$  between $0.15-0.3m$ for the scenes in 7Scenes dataset. 

\noindent$\textbf{DFNet}_{dm}$. $\text{DFNet}_{dm}$~\cite{chen2022dfnet} (DFNet with feature-metric direct matching) is the SOTA model in semi-supervised learning  APR. We integrate the official pre-trained model\footnote{\url{https://github.com/ActiveVisionLab/DFNet}}
%that can reproduce the result of the paper~\cite{chen2022dfnet} 
and apply it to 7Scenes dataset and Cambridge dataset. 
%We set $\gamma$ between 20 to 40 and  $d_{th}$  between $0.15-0.3m$ for the 7Scenes dataset. We set $\gamma$ between 15 to 30 and  $d_{th}$  between $1-2m$ for the Cambridge dataset. 

\noindent\textbf{MS-T}. MS-Transformer~\cite{shavit2021learning} (MS-T) is the SOTA model in supervised learning APR, which aggregates the activation maps with self-attention and queries the scene-specific information.
We integrate the official pre-trained model\footnote{\url{https://github.com/yolish/multi-scene-pose-transformer}} 
in our pipeline and apply it to the Cambridge dataset. %We set $\gamma$ between 15 to 30, and  $d_{th}$  between $1-2m$ for each scene.  

\textbf{We note APR methods integrated into our keyframe selection pipeline as $\text{APR}^{ks}$}. 
Since some methods do not provide pre-trained models for all datasets, or the pre-trained model cannot reproduce the original paper's results, we omit MS-T in 7Scenes and DFNet in Cambridge.
We set the match number threshold $\gamma$ between 20 to 40, and the distance threshold  $d_{th}$  between $0.15-0.3m$ for the scenes in 7Scenes dataset. We set $\gamma$ between 15 to 30 and  $d_{th}$  between $1-2m$ for the Cambridge dataset. We provide the specific hyperparameters for each model and dataset as supplementary material.
%\lck{Our pipeline is actually not very sensitive to hyperparameters, and is able to improve the performance of the original APR using hyperparameters within a certain range.}
%%
\subsection{Evaluation on the 7Scenes Dataset}

We compare our pipeline with labeled methods of single-frame APR, unlabeled methods, uncertainty-aware APR methods and sequential-based APR over the 7Scenes dataset in Tables~\ref{tab:df_7s_rank} and~\ref{tab:dm_7s_rank}. Table~\ref{tab:df_7s_rank} also compares the current baseline of IR-based visual positioning (DenseVLAD+Inter.)~\cite{sattler2019understanding}. $\text{DFNet}^{ks}$ ranks first in both median position and orientation errors over scene average. Our pipeline improves the accuracy of $\text{DFNet}$ as much as $22.2\%$ on position error and $17.7\%$ on orientation error. Similarly, $\text{DFNet}_{dm}^{ks}$ ranks first in median error over scene average in Table~\ref{tab:dm_7s_rank}.  
Our pipeline improves the accuracy of $\text{DFNet}_{dm}$ as much as $28.6\%$ on position error and $22\%$ on orientation error.

The prediction accuracy of almost all the APR methods containing uncertainty estimation are lower than SOTA APR methods (DFNet and DFNet$_{dm}$).
Only CoordiNet+LENS, which augments the training set with a large number of synthetic images, achieves an accuracy close to our method. With \sysname applied to the SOTA APR DFNet$_{dm}$, we can achieve much higher accuracy than APR methods that output uncertainty estimation.
Since we perform uncertainty estimation at test time, any APR method can be easily integrated into KS-APR, reflecting our approach's flexibility compared to designing the uncertainty estimation on the architecture of neural networks and loss function during training.

Table~\ref{tab:dfdm_7s_level} displays how our proposed method improves the accuracy of $\text{DFNet}$ and $\text{DFNet}_{dm}$. We display the proportion of images predicted with a high (0.25$m$, $2^{\circ}$), medium (0.5$m$, $5^{\circ}$), and low (5$m$, $10^{\circ}$) accuracy, as followed by Sattler \emph{et al.}~\cite{sattler2018benchmarking}. The keyframe selection method applied to $\text{DFNet}^{ks}$ preserves between 60\% and 72\% of the test images and improves the proportion of images detected at all accuracy levels. Similarly, $\text{DFNet}_{dm}^{ks}$ keeps 57\% to 70\% of the test images as keyframes and achieves higher percentages in all precision levels. In three datasets, it localizes all images with an accuracy of at least 5$m$ and $10^{\circ}$.

\subsection{Evaluation on Cambridge Dataset}

\begin{table*}[t]
\caption{\textbf{Labeled} training methods of \textbf{single-frame} APR, \textbf{IR} baseline (DenseVLAD+Inter.) and \textbf{classical} APR methods with \textbf{uncertainty estimation} (\text{Bayesian PN and AD-PN}) compared with our method on Cambridge dataset. We report the median position/orientation errors in m/° and the respective rankings over scene average as in~\cite{shavit2021learning,chen2022dfnet}. Note that our pipeline achieves top performance by only keeping keyframe images. }%The percentage of keyframes is detailed in Table~\ref{tab:transdm_cam_level}.}
$$
\begin{array}{l|cccc|c|c}
\hline \text{Labeled Methods \& \text{IR}} & \text{Kings } & \text{Hospital } & \text{Shop } & \text{Church } &\text{Average } & \text{Ranks }  \\
\hline \text{PoseNet(PN)\cite{kendall2015posenet} } & 1.66 / 4.86 & 2.62 / 4.90 & 1.41 / 7.18 & 2.45 / 7.96 & 2.04 / 6.23 & 10/9 \\
\text{PN Learn} \sigma^2 \text{\cite{kendall2017geometric}} & 0.99 / 1.06 & 2.17 / 2.94 & 1.05 / 3.97 & 1.49 / 3.43 & 1.43 / 2.85 & 5/3 \\
\text{geo. PN\cite{kendall2017geometric} } & 0.88 / \mathbf{1.04 }& 3.20 / 3.29 & 0.88 / 3.78 & 1.57 / 3.32 & 1.63 / 2.86 & 6/4\\
\text{LSTM PN\cite{walch2017image} } & 0.99 / 3.65 & 1.51/ 4.29 & 1.18 / 7.44 & 1.52 / 6.68 & 1.30 / 5.51 & 4/8\\
\text{MapNet\cite{brahmbhatt2018geometry} } & 1.07 / 1.89 & 1.94 / 3.91 & 1.49 / 4.22 & 2.00 / 4.53 & 1.63 / 3.64 & 6/6 \\
\text{DFNet \cite{chen2022dfnet} } & \mathbf{0.73} / 2.37 & 2.00 / 2.98 & \mathbf{0.67 / 2.21} & 1.37 / 4.03 & 1.19 / 2.90 & 2/5 \\
\text{MS-T\cite{shavit2021learning}} & 0.83 / 1.47 & 1.81 / 2.39 & 0.86 / 3.07 & 1.62 / 3.99 & 1.28 / 2.73 & 3/2\\
\hline
\text {Bayesian PN\cite{kendall2016modelling} } & 1.74/ 4.06 &2.57/ 5.14&1.25/ 7.54  &2.11/ 8.38& 1.92/6.28& 9/10\\
\text{AD-PN\cite{huang2019prior}} & 1.3/1.67 & 2.28/4.80& 1.22/6.17& -/-& 1.60/4.21& -/- \\\hline
\text{DenseVLAD+Inter.\cite{sattler2019understanding}} & 1.48/4.45&2.68/4.63&0.90/4.32&1.62/6.06& 1.67/4.87& 8/7 \\ \hline
\text{MS-T}^{ks}\text{(ours)} &0.78/1.43& \mathbf{1.27/1.81} & 0.84/3.02  & \mathbf{1.34}/3.41& \mathbf{1.02/2.49} & \mathbf{1/1}\\
\hline
\end{array}
$$
\label{tab:trans_cam_rank}
\end{table*}

\begin{table*}[t]
\caption{\textbf{Unlabeled} methods ($\text{DFNet}_{dm}$), \textbf{sequential-based} APR (VLocNet) and the \textbf{latest} APR methods with \textbf{uncertainty estimation} (\text{CoordiNet, CoordiNet+LENS, BMDN, VaPoR}) compared with our method on Cambridge dataset. We report the median position/orientation errors in m/° and $\text{DFNet}_{d m}^{ks}$ achieves top performance by only keeping keyframes. The percentage of keyframes is detailed in Table~\ref{tab:transdm_cam_level}.
We omit prior APR methods which did not publish results in Cambridge.}
$$
\begin{array}{l|c|ccccc|c}
\hline
 & \text{Seq.  APR} &  &  \text{1-frame APR}  &&  &  & \text{Filter}    \\ \hline
\text{Dataset} & \text{VLocNet\cite{valada2018deep}} & \text{CoordiNet\cite{moreau2022coordinet}} & \text{CoordiNet+LENS\cite{moreau2022lens}} &\text{BMDN\cite{bui20206d}} & \text{VaPoR\cite{zangeneh2023probabilistic}}&\text{DFNet}_{d m}\text{\cite{chen2022dfnet}}  & \text{DFNet}_{d m}^{ks}\text{(ours)}    \\ \hline
\text{Kings}     & 0.84/1.42  &  0.70/2.92 & \mathbf{0.33}/\mathbf{0.5} &0.88/1.04 &  1.65/2.88 & 0.43/0.87 & 0.43/0.72  \\
\text{Hospital}  & 1.08/2.41  & 0.97/2.08  & 0.44/0.9 & 3.2/3.29& 2.06/4.33  & 0.46/0.87  & \mathbf{0.36}/\mathbf{0.71}   \\
\text{Shop}       & 0.59/3.53   & 0.69/3.74   & 0.27/1.6 & 0.88/3.78& 1.02/6.03 & 0.16/0.59 & \mathbf{0.12}/\mathbf{0.54} \\
\text{Church}     & 0.63/3.91 & 1.32/3.56   & 0.53/1.6  &1.57/3.32 &  1.80/5.90 & 0.50/1.49 & \mathbf{0.44/1.28}  \\ \hline
\text{Average} & 0.78/2.82 & 0.92/2.58 & 0.39/1.15 & 1.63/2.86 & 1.63/4.79 & 0.39/0.96&\mathbf{0.34/0.81}\\\hline
\end{array}
$$
\label{tab:dm_cam_rank}
\end{table*}

\begin{table*}[!t]
\caption{Percentage of images with pose predicted with high (0.25m, $2^{\circ}$), medium (0.5m, $5^{\circ}$), and low (5m, $10^{\circ}$) accuracy~\cite{sattler2018benchmarking}. The value in parentheses represents the percentage of keyframes identified in the test set by our method.}
$$
\begin{array}{l|cc||cc}
\hline
\text{Dataset}&\text{MS-T\cite{shavit2021learning}} & \text{MS-T}^{ks}\text{(ours)}& \text{DFNet}_{d m}\text{\cite{chen2022dfnet}} & \text{DFNet}_{d m}^{ks}\text{(ours) }   \\ \hline
\text{Kings}   &3.5/20.7/97.1 & \mathbf{3.7/21.2/98.8}(70\%)& 20.6/58.0/96.2 & \mathbf{22.6/59.5/99.2}(73\%)  \\
\text{Hospital} &2.2/7.1/84.1 & \mathbf{6.5/17.3/93.4}(25\%) & 21.4/53.8/\mathbf{100}  & \mathbf{31.0/64.0/100}(41\%) \\
\text{Shop}    &4.8/21.4/93.2 & \mathbf{5.3/23.1/94.7}(92\%)  & 73.8/88.3/98.1 & \mathbf{81.9/97.5/100}(81\%)\\
\text{Church} & 0/2.3/83.8& \mathbf{0/3.0/91.8}(62\%) & 19.6/49.8/96.2 & \mathbf{22.1/56.2/98.9 }(74\%)\\ \hline
\end{array}
$$
\label{tab:transdm_cam_level}
\end{table*}

We compare labeled methods of single-frame APR, unlabeled methods, uncertainty-aware APR and sequential-based APR in Tables~\ref{tab:trans_cam_rank} and~\ref{tab:dm_cam_rank}. Table~\ref{tab:trans_cam_rank} also compares the current baseline of IR-based visual positioning (DenseVLAD+Inter.). The methods presented in Table~\ref{tab:trans_cam_rank} perform differently depending on the scene in the Cambridge Dataset, whether in terms of position or orientation. However,  $ \text{MS-T}^{ks}$ still ranks 1/1 in median error over scene average, while MS-T  ranks 3/2.
In terms of the sequential APR and  APR methods with latest uncertainty estimation presented in Table~\ref{tab:dm_cam_rank}, 1-frame $\text{DFNet}_{dm}^{ks}$ with keyframes selection surpasses all previous methods in median error over scene average. Only CoordiNet+LENS present a slightly higher accuracy in KingsCollege scene. 

Table~\ref{tab:transdm_cam_level} displays how our keyframe selection method improves the accuracy of $\text{MS-T}$ and $\text{DFNet}_{dm}$. \sysname keeps 27\% to 92\% of the test images in the case of $\text{MS-T}$, and 41\% to 81\% of the test images for $\text{DFNet}_{dm}$. \sysname also improves the percentage of images in each accuracy for both methods. Similar to the 7Scenes dataset, $\text{DFNet}_{dm}^{ks}$ can accurately predict all the keyframes of two scenes with an accuracy of (5$m$, $10^{\circ}$).

\subsection{Analysis}
\label{subsec:analy}
\begin{figure}[hbt!]
  \centering
  \includegraphics[width=.9\linewidth]{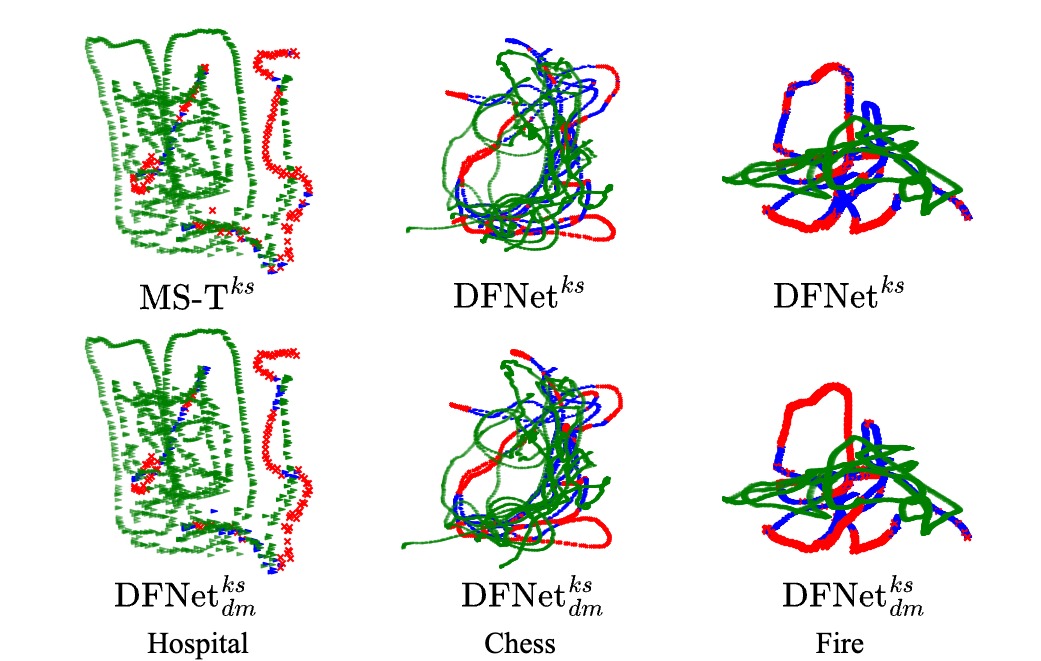}
  \caption{ %For Hospital, the upper one is the result of $\text{MS-T}^{ks}$ and the below one is $\text{DFNet}_{dm}^{ks}$. For Chess and Fire,  the upper one is $\text{DFNet}^{ks}$ and the below one is $\text{DFNet}_{dm}^{ks}$. 
  Camera ground truth trajectory. Green: training set; Blue: keyframes in testset; Red: filtered images in testset. Our proposed method primarily identifies frames close to the training set as keyframes for improving accuracy.}
\label{fig:route}
\end{figure}

\begin{figure}[!t]
 \centering
  \subfigure[Chess ($\text{DFNet}_{dm}^{ks}$)]{
 \label{fig:subfig:f} %% label for second subfigure
 \includegraphics[height=1.0in,width=1.4in]{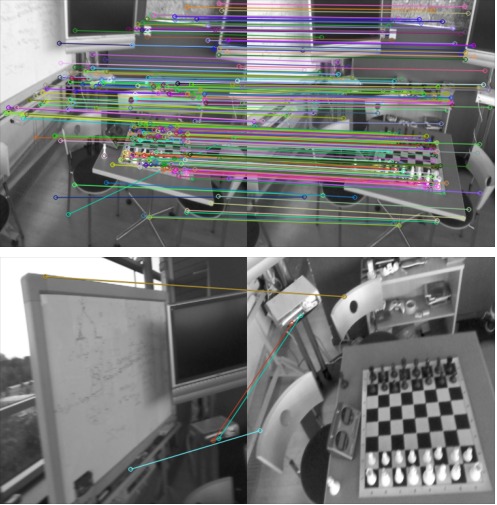}}
 \subfigure[Hospital ($\text{DFNet}_{dm}^{ks}$)]{
 \label{fig:subfig:a} %% label for first subfigure
 \includegraphics[height=1.0in,width=1.4in]{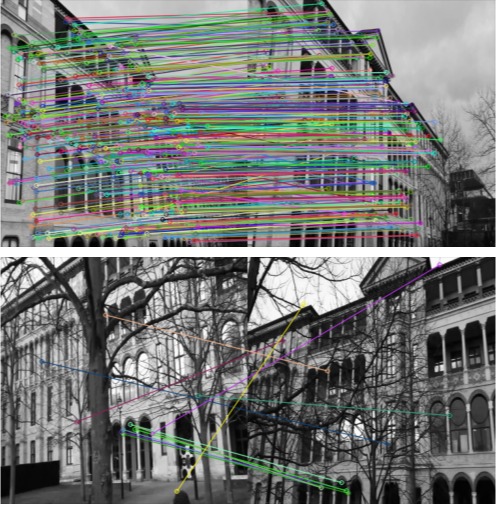}}
 \hspace{0.1in}
  \subfigure[Pumpkin ($\text{DFNet}^{ks}$)]{
 \label{fig:subfig:f} %% label for second subfigure
 \includegraphics[height=1.0in,width=1.4in]{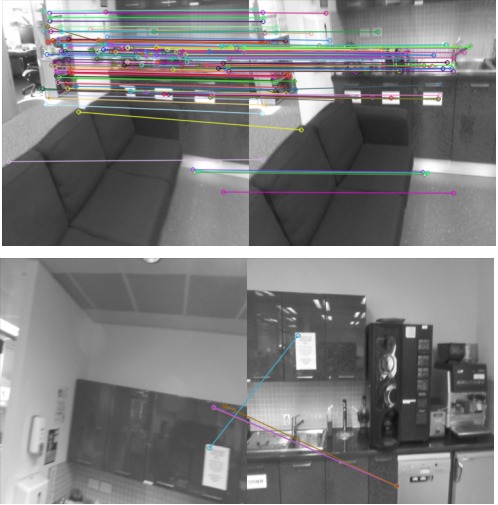}}
   \subfigure[Hospital ($\text{MS-T}^{ks}$)]{
 \label{fig:subfig:f} %% label for second subfigure
 \includegraphics[height=1.0in,width=1.4in]{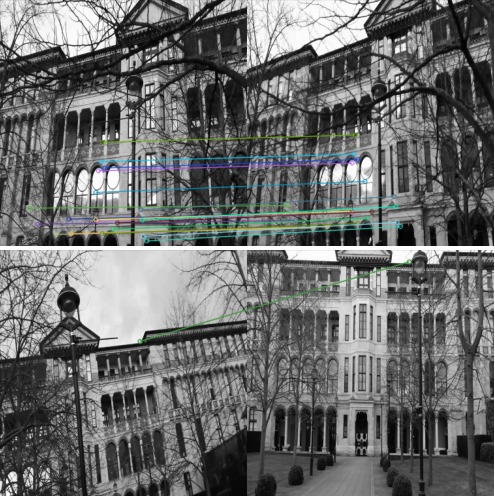}}
 \caption{Examples of keyframes and removed images in Cambridge and 7Scenes. For each pair, left side is the query image in test set, right side is the image retrieved by Algorithm~\ref{alg:retrieval} using pretrained APR. The upper row of each subfigure shows a query image in test set selected as a keyframe, and the lower row is a query image featuring too little matches to be considered a keyframe.}
 \label{fig:good_bad} %% label for entire figure
\end{figure}

We visualize the identified keyframes and the corresponding database images to analyze how our pipeline improves the robustness and accuracy of APR methods. As shown in Figure~\ref{fig:good_bad}, the keyframes identified by $\text{MS-T}^{ks}$ and $\text{DFNet}_{dm}^{ks}$ in the Hospital dataset have similar images that meet the requirement of our pipeline. The removed images are consistent with the principles described in Section~\ref{subsec:imgretrieval}: rejected images either are too far from images in the training set or present a significant orientation disparity at closer positions, making it difficult to get enough feature matches. Similarly, in the 7Scenes dataset, the keyframes and removed images that DFNet and $\text{DFNet}_{dm}$ can identify also fit such characteristics. The correlation between predicted uncertainty and the actual pose error is not rigorous matching in \cite{kendall2016modelling,deng2022deep,bui20206d}. 
These systems only output the uncertainty of the APR pose estimation without determining the cause for such uncertainty.
%When we get the uncertainty of the output from an APR, we cannot determine why the prediction of this query is the way it is. 
Our method shows the reliability of pose prediction in a more intuitive and clearer way. One can get a better understanding of exactly which position and orientation distributions the model lacks to learn. In 7-scenes dataset, DFNet$_{dm}^{ks}$ is almost comparable to popular 3D methods as shown in Table~\ref{tab:3d_7s}. 
However, in terms of computation and storage overhead, our method is far better than the 3D methods.
 
The Hospital scene in the Cambridge dataset has many repetitive and symmetric features (see Figure~\ref{fig:good_bad}). As such, we set the ratio test to 0.5 instead of 0.7 to get good matches. 

Figure~\ref{fig:route} illustrates the ground truth camera pose trajectories. Training set images are represented in green. Query paths are composed of keyframes (blue) and filtered images (red).
The Hospital dataset is filtered by $\text{MS-T}^{ks}$ (upper) and $\text{DFNet}_{dm}^{ks}$ (below), while Chess and Fire datasets are filtered by $\text{DFNet}^{ks}$ (upper) and $\text{DFNet}_{dm}^{ks}$ (below). In each subfigure, most filtered images are located far away from the places where the training images are taken. In the Hospital scene of the Cambridge dataset, the distribution of the test set has a large bias compared with the training set, which leads to fewer keyframes identified by $\text{MS-T}^{ks}$ and $\text{DFNet}_{dm}^{ks}$, explaining the large percentage of images deletion in Table~\ref{tab:transdm_cam_level}. In the Chess and Fire scenes, the selected keyframes are less separated from the bad images by $\text{DFNet}^{ks}$ than the ones by $\text{DFNet}_{dm}^{ks}$, which matches the difference of performance presented in Table~\ref{tab:dfdm_7s_level}. This visualization confirms that our method can identify keyframes close to the training set, providing more robust and reliable results.

%\subsection{Influence of Hyperparameters}
Table~\ref{tab:paraset} displays the improvement brought by our pipeline over $\text{DFNet}_{dm}$ for different values of $d_{th}$ and $\gamma$ in median error and all precision levels. Although the highest accuracy is reached for values of $d_{th}$ and $\gamma$ that result in stricter keyframe selection, differences in accuracy remain small, showing our pipeline's low sensitivity to the hyperparameters. More images and tables can be seen in the supplementary material.

%Although notable, the difference in accuracy for different  $d_{th}$ and $\gamma$ remains small. More tables are in the supplementary material.

\subsection{System efficiency}
We evaluate the processing time of the proposed pipeline on a PC equipped with an Intel Core i9-12900K CPU (5.20GHz) CPU, 32GB of RAM, and an NVIDIA GeForce GTX 3090 GPU. Based on the implementation in Section~\ref{subsec:imp},  
SIFT feature detection and extraction (CPU implementation) take $8-10ms$, while feature matching takes $1-2ms$. As our method performs feature matching for a single image, the processing time of the pipeline is significantly reduced. We test the image retrieval performance on King's College dataset with 1220 training images and their ground truth pose in the database. Retrieval takes $1-2ms$ with a $\mathcal{O}(n)$ Algorithm~\ref{alg:retrieval}. Overall, our pipeline only adds less than $15ms$ to the total inference time of APRs.

We also evaluate the average processing time of the proposed pipeline on an iPhone 14 Pro Max device with 200 samples. We transferred pre-trained PoseNet using ResNet34 as backbone into ONNX format and integrated it into a Unity application. The average processing time per image using
OpenCVforUnity\footnote{\url{https://assetstore.unity.com/packages/tools/integration/opencv-for-unity-21088}} is $37 ms$.  PoseNet takes $39.5 ms$ to infer an image. The time of pose-based image retrieval is negligible, less than $1ms$. SIFT feature detection and extraction using OpenCVforUnity takes $36ms$, while feature matching takes $4ms$. Therefore, our pipeline only adds less than $40ms$ to the total inference time of APRs, even on mobile devices. Besides, \sysname does not need point cloud models like 3D methods and databases for original images and 2d-3d correspondences. \sysname only requires a small database for low-resolution resized images with ground truth labels and a pre-trained APR model. 
%making it suitable for embedded and mobile applications for real-time localization. 

\section{Conclusion}

This paper introduces \sysname, an APR-agnostic pipeline that relies on a new featureless IR method to identify the most similar images in the training set to the query image, followed by local feature extraction and matching to assess the reliability of the estimated pose. 
We evaluate three types of APR models on indoor and outdoor datasets. Our method outperforms current single-image  APR methods by as much as 28.6\% on position error and 22\% on orientation error, and minimizes large-error estimations with only a 15\,$ms$ overhead on the PC. Even if this pipeline runs completely on the smartphone, it can be finished within 120$ms$.
It also enables $\text{DFNet}_{dm}$ to achieve higher accuracy than sequential APR methods. Our approach provides a new state-of-the-art localization pipeline to enable APR methods in real-life applications.

% In this paper,  we introduce an  APR-agnostic pipeline implementing a new similarity-based IR method to fetch the most similar images in the training set, followed by local features extraction and matching to identify reliable images and discard bad ones. 
% % measurement method to determine the similarity between query images and images in the training set. We design a feature-less IR algorithm using the APR 6DoF result and the new similarity measurement method.   We design an APR-agnostic pipeline that implements new IR method to fetch images that are similar to the query image, followed by local features extraction and matching to identify reliable images and discard bad ones. 
% \iffalse We evaluate three types of APR models separately on indoor and outdoor datasets and find that the median error on both position and orientation is reduced for all models incorporating our pipeline, and the ratio of estimations with large error is significantly lower across datasets.\fi
% We evaluate three types of APR models on indoor and outdoor datasets. Our method outperforms both current SOTA single-image and sequential APR methods by as much as 28.6\% on position error and 22\% on orientation error, with large-error estimations significantly reduced. Our approach provides a new state-of-the-art localization pipeline to enable APR methods in real-life applications.

%\bibliographystyle{abbrv}
\bibliographystyle{abbrv-doi}

\bibliography{template}
\end{document}